\documentclass[11pt]{article}
\usepackage{acl2015}
\usepackage{times}
\usepackage{url}
\usepackage{latexsym}

\usepackage{times}
\usepackage{amsmath}
\usepackage{amsthm}
\usepackage{amssymb}
\usepackage{graphicx}
\usepackage{xspace}
\usepackage{tabularx}
\usepackage{multicol}
\usepackage{multirow}
\usepackage{url}
\usepackage{wrapfig}
\usepackage{comment}
\usepackage{listings}
\usepackage{color}
\usepackage[utf8]{inputenc}
\usepackage{fancyvrb}
\usepackage{changes}
\usepackage{booktabs}
\usepackage{color}
\usepackage[normalem]{ulem}

\definechangesauthor[color=blue]{caglar}
\definechangesauthor[color=green]{orf}

\newcommand{\f}{\text{f}}

\newcommand{\bmx}[0]{\begin{bmatrix}}
\newcommand{\emx}[0]{\end{bmatrix}}

\newcommand{\vect}[1]{\mathbf{#1}}
\newcommand{\vects}[1]{\boldsymbol{#1}}
\newcommand{\matr}[1]{\mathbf{#1}}

\newcommand{\vb}[0]{\vect{b}}
\newcommand{\vc}[0]{\vect{c}}

\newcommand{\vh}[0]{\vect{h}}
\newcommand{\vv}[0]{\vect{v}}
\newcommand{\vx}[0]{\vect{x}}

\newcommand{\vs}[0]{\vect{s}}

\newcommand{\vy}[0]{\vect{y}}
\newcommand{\vg}[0]{\vect{g}}

\newcommand{\mW}[0]{\matr{W}}

\newcommand{\TT}[0]{\vects{\theta}}

\newcommand{\E}[0]{\mathbb{E}}

\newcommand{\ola}{\overleftarrow}
\newcommand{\ora}{\overrightarrow}

\graphicspath{ {./figures/} }

\title{On Using Monolingual Corpora in Neural Machine Translation}

\author{Caglar Gulcehre$^\star$ \\
  Universit\'e de Montr\'eal \\\And
  Orhan Firat$^{\star,\dagger}$ \\
  Middle East Technical University \\\And
  Kelvin Xu\\
  Universit\'e de Montr\'eal \\\AND
  Kyunghyun Cho\\
  Universit\'e de Montr\'eal \\\And
  Lo\"ic Barrault\\
  Universit\'e du Maine \\\And
  Huei-Chi Lin\\
  Universit\'e du Maine\\\And
  Fethi Bougares\\
  Universit\'e du Maine\\\AND
  Holger Schwenk\\
  Universit\'e du Maine\\\And
  Yoshua Bengio\\
  Universit\'e de Montr\'eal\\
  CIFAR Senior Fellow\\ \\}
\date{}

\begin{document}
\maketitle
\let\thefootnote\relax\footnote{$^\star$ Equal contribution. Order has been determined with a coin flip.}
\let\thefootnote\relax\footnote{$^\dagger$ Work done while author was at Universit\'e de Montr\'eal}

\vspace{2.5em}

\begin{abstract}

Recent work on end-to-end neural network-based architectures for machine
translation has shown promising results for En-Fr and En-De translation.
Arguably, one of the major factors behind this success has been the availability of high
quality parallel corpora. In this work, we investigate how to leverage abundant
monolingual corpora for neural machine translation. Compared to a phrase-based
and hierarchical baseline, we obtain up to $1.96$ BLEU improvement on the
low-resource language pair Turkish-English, and $1.59$ BLEU on the focused domain
task of Chinese-English chat messages. While our method was initially targeted
toward such tasks with less parallel data, we show that it also extends to high
resource languages such as Cs-En and De-En where we obtain an improvement of
$0.39$ and $0.47$ BLEU scores over the neural machine translation baselines,
respectively.

\end{abstract}

\section{Introduction}
\label{sec:intro}
{\it Neural machine translation} (NMT) is a novel approach to machine
translation that has shown promising results
\cite{Kalchbrenner2013,Sutskever2014,Cho2014,bahdanau2014neural}. Until
recently, the application of neural networks to machine translation was
restricted to extending standard machine translation tools for rescoring
translation hypotheses or re-ranking n-best lists (see, e.g.,
\cite{Schwenk2012,Schwenk2007}. In contrast, it has been shown that, it is
possible to build a competitive translation system for English-French and
English-German using an end-to-end neural network architecture~\cite{Sutskever2014,Jean2014}
(also see Sec.~\ref{sec:background}). 

Arguably, a large part of the recent success of these methods has been due to
the availability of large amounts of high quality, sentence aligned corpora. In
the case of low resource language pairs or in a task with heavy domain
restrictions, there can be a lack of such sentence aligned corpora. In
contrast, monolingual corpora is almost always universally available. Despite
being ``unlabeled'', monolingual corpora still exhibit rich linguistic
structure that may be useful for translation tasks. This presents an
opportunity to leverage such corpora to give hints to an NMT system.

In this work, we present a way to effectively integrate a language model (LM)
trained only on monolingual data (target language) into an NMT system.  We
provide experimental results that incorporating monolingual corpora can improve
a translation system on a low-resource language pair (Turkish-English) and a
domain restricted translation problem (Chinese-English SMS chat). In addition,
we show that these methods improve the performance on the relatively
high-resource German-English (De-En) and Czech-English (Cs-En) translation
tasks.

%

In the following section (Sec.~\ref{sec:background}), we review recent work
in neural machine translation. We present our basic model architecture in
Sec.~\ref{sec:model_des}~ and describe our shallow and deep fusion approaches
in Sec.~\ref{sec:nmt_lm}. Next, we describe our datasets in~Sec.~\ref{sect:dataset}. 
Finally, we describe our main experimental results in Sec.~\ref{sec:experiments}.

\section{Background: Neural Machine Translation}
\label{sec:background} 

Statistical machine translation (SMT) systems maximize the conditional probability 
$p(\vy \mid \vx)$ of a correct target translation $\vy$ given a source sentence $\vx$. 
This is done by maximizing separately a language model $p(\vy)$ and the (inverse) translation
model $p(\vx \mid \vy)$ component by using Bayes' rule:
\begin{align*}
    p(\vy \mid \vx) \propto p(\vx \mid \vy) p(\vy).
\end{align*}

This decomposition into a language model and translation model is meant to make full
use of available corpora: monolingual corpora for fitting the language model and
parallel corpora for the translation model. In reality, however, SMT systems
tend to model $\log p(\vy\mid\vx)$ directly by linearly combining multiple features
by using a so-called log-linear model:
\begin{align}
    \label{eq:trad_smt}
    \log p(\vy \mid \vx) = \sum_{j} f_j(\vx, \vy) + C,
\end{align}
where $f_j$ is the $j$-th feature based on both or either of the source and
target sentences, and $C$ is a normalization constant which is often ignored.
These features include, for instance, pair-wise statistics between two
sentences/phrases. The log-linear model is fitted to data, in most cases, by
maximizing an automatic evaluation metric other than an actual conditional
probability, such as BLEU.

Neural machine translation, on the other hand, aims at directly optimizing
$\log p(\vy \mid \vx)$ including the feature extraction as well as the
normalization constant by a single neural network.  This is typically done under
the encoder-decoder framework~\cite{Kalchbrenner2013,Cho2014,Sutskever2014}
consisting of neural networks. The first network encodes the
source sentence $\vx$ into a continuous-space representation from which the
decoder produces the target translation sentence. By using RNN 
architectures equipped to learn long term dependencies such as Gated Recurrent
Units (GRU) or Long Short-Term Memory (LSTM), the whole system can be trained
end-to-end~\cite{Cho2014,Sutskever2014}.

Once the model learns the conditional distribution or translation model, given
a source sentence we can find a translation that approximately maximizes the
conditional probability using, for instance, a beam search algorithm.

\section{Model Description}
\label{sec:model_des}

We use the model recently proposed by \cite{bahdanau2014neural}
that learns to jointly (soft-)align and translate as the baseline neural machine
translation system in this paper. Here we describe in detail this model to
which we refer as ``NMT''.

The encoder of the NMT is a bidirectional RNN which consists of forward and
backward RNNs~\cite{Schuster1997}. The forward RNN reads the input
sequence/sentence $\vx=(x_1, \dots, x_T)$ in a forward direction, resulting in a
sequence of hidden states $(\ora{\vh}_1, \dots, \ora{\vh}_T)$. The backward RNN
reads $\vx$ in an opposite direction and outputs $(\ola{\vh}_1, \dots,
\ola{\vh}_T)$. We concatenate a pair of hidden states at each time step to build
a sequence of {\it annotation} vectors $(\vh_1, \dots, \vh_T)$, where 
\[
    \vh_j^\top = \left[ 
            \ola{\vh}_j^\top ; \ora{\vh}_j^\top
    \right].
\]
Each annotation vector $\vh_j$ encodes information about the $j$-th word with
respect to all the other surrounding words in the sentence.

In our decoder, which we construct with a single layer RNN, at each timestep
$t$ a soft-alignment mechanism first decides on which annotation vectors are
most relevant. The relevance weight $\alpha_{tj}$ of the $j$-th annotation
vector for the $t$-th target word is computed by a feedforward neural network
$\f$ that takes as input $\vh_j$, the previous decoder's hidden state
$\vs_{t-1}$ and the previous output $\vy_{t-1}$:
\begin{align*}
    e_{tj} = \f(\vs_{t-1}, \vh_j, \vy_{t-1}).
\end{align*}
The outputs $e_{tj}$ are normalized over the sequence of the annotation vectors
so that the they sum to $1$:
\begin{equation}
    \label{eq:dec:alpha}
    \alpha_{tj} = \frac{\text{exp}(e_{tj})}{\sum_{k=1}^T\text{exp}(e_{tk})},
\end{equation}
and we call $\alpha_{tj}$ a relevance score, or an alignment weight, of the
$j$-th annotation vector.

The relevance scores are used to get the {\it context vector} $\vc_t$ of the
$t$-th word in the translation:
\begin{align*}
    \vc_t = \sum_{j=1}^T \alpha_{tj} \vh_j ~,
\end{align*}

Then, the decoder's hidden state $\vs_t^{\text{TM}}$ at time $t$ is computed
based on the previous hidden state $\vs_{t-1}^{\text{TM}}$, the context vector
$\vc_t$ and the previously translated word $\vy_{t-1}$:
\begin{equation}
    \label{eq:dec:hidden}
    \vs_t^{\text{TM}} = \f_r(\vs_{t-1}^{\text{TM}}, \vy_{t-1}, \vc_t),
\end{equation}
where $\f_r$ is the gated recurrent unit~\cite{Cho2014}.

We use a deep output layer~\cite{Pascanu2014rec} to compute the conditional
distribution over words: 
\begin{equation}
    \begin{split}
    \label{eq:dec:output}
    p(&\vy_t | \vy_{<t}, \vx) \propto \\
    & \exp(\vy_t^{\top}(\mW_o 
    \f_o(\vs_t^{\text{TM}}, \vy_{t-1}, \vc_t) + \vb_o)),
    \end{split}
\end{equation}
where $\vy_t$ is a one-hot encoded vector indicating one of the words in the
target vocabulary. $\mW_o$ is a learned weight matrix and $\vb_o$ is a bias.
$\f_o$ is a single-layer feedforward neural network with a two-way maxout
non-linearity~\cite{Goodfellow2013}.

The whole model, including both the encoder and decoder, is jointly trained to
maximize the (conditional) log-likelihood of the bilingual training corpus:
\begin{align*}
    \max_{\TT} \frac{1}{N} \sum_{n=1}^N \log p_{\theta}(\vy^{(n)} | \vx^{(n)}),
\end{align*}
where the training corpus is a set of $(\vx^{(n)}, \vy^{(n)})$'s, and $\TT$
denotes a set of all the tunable parameters.

\section{Integrating Language Model into the Decoder} 
\label{sec:nmt_lm}

In this paper, we propose two alternatives to integrating a language model into
a neural machine translation system which we refer as {\it shallow fusion}
(Sec.~\ref{sec:shallow}) and {\it deep fusion} (Sec.~\ref{sec:deep}). Without
loss of generality, we use a language model based on recurrent neural networks
(RNNLM, \cite{mikolov2011rnnlm}) which is equivalent to the decoder described in
the previous section except that it is not biased by a context vector (i.e.,
$\vc_t = 0$ in Eqs.~\eqref{eq:dec:hidden}--\eqref{eq:dec:output}). 

In the sections that follow, we assume that both an NMT model (on parallel
corpora) as well as a recurrent neural network language model (RNNLM, on larger 
monolingual corpora) have been pre-trained separately before being integrated.  
We denoted the hidden state at time $t$ of the RNNLM with $\vs_t^{\text{LM}}$.

\subsection{Shallow Fusion}
\label{sec:shallow}

Shallow fusion is analogous to how language models are used in the decoder of a
usual SMT system~\cite{koehn2010}. At each time step, the translation model
proposes a set of candidate words. The candidates are then scored according to
the weighted sum of the scores given by the translation model and the language
model. 

More specifically, at each time step $t$, the translation model (in this case,
the NMT) computes the score of every possible next word for each hypothesis all
of hypotheses $\left\{ \vy_{\leq t-1}^{(i)}\right\}$. Each score is the summation of the score of the
hypothesis and the score given by the NMT to the next word. All these new
hypotheses (a hypothesis from the previous timestep with a next word appended
at the end) are then sorted according to their respective scores, and the top
$K$ ones are selected as candidates $\left\{ \hat{\vy}^{(i)}_{\leq t}
\right\}_{i=1,\dots,K}$.

We then rescore these hypotheses with the weighted sum of the scores by the NMT
and RNNLM, where we only need to recompute the score of the ``new word'' at the
end of each candidate hypothesis. The score of the new word is computed by
\begin{equation}
    \label{eq:dec:output_shallowfusion}
    \begin{split}
        \log p (\vy_t=k)& = \log p_{\text{TM}} (\vy_t = k) \\ 
        & + 
        \beta \log p_{\text{LM}} (\vy_t = k),
    \end{split}
\end{equation}
where $\beta$ is a hyper-parameter that needs to be tuned to maximize the
translation performance on a development set.

\begin{figure*}[th]
\begin{minipage}{0.48\textwidth}
        \centering
            \includegraphics[width=0.9\columnwidth]{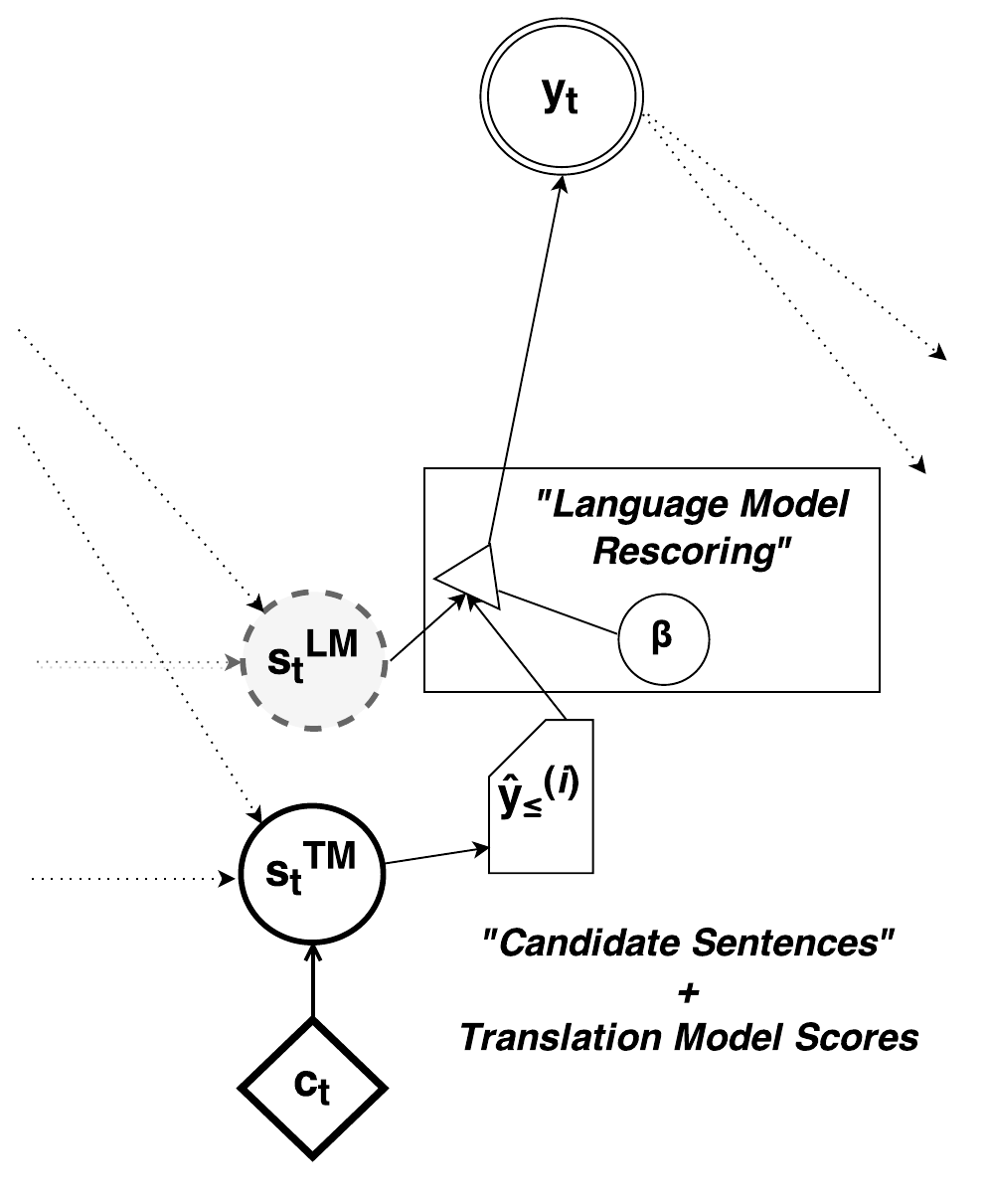}
\end{minipage}
\hspace{2mm}
    \begin{minipage}{0.48\textwidth}
        \centering
            \includegraphics[width=\columnwidth]{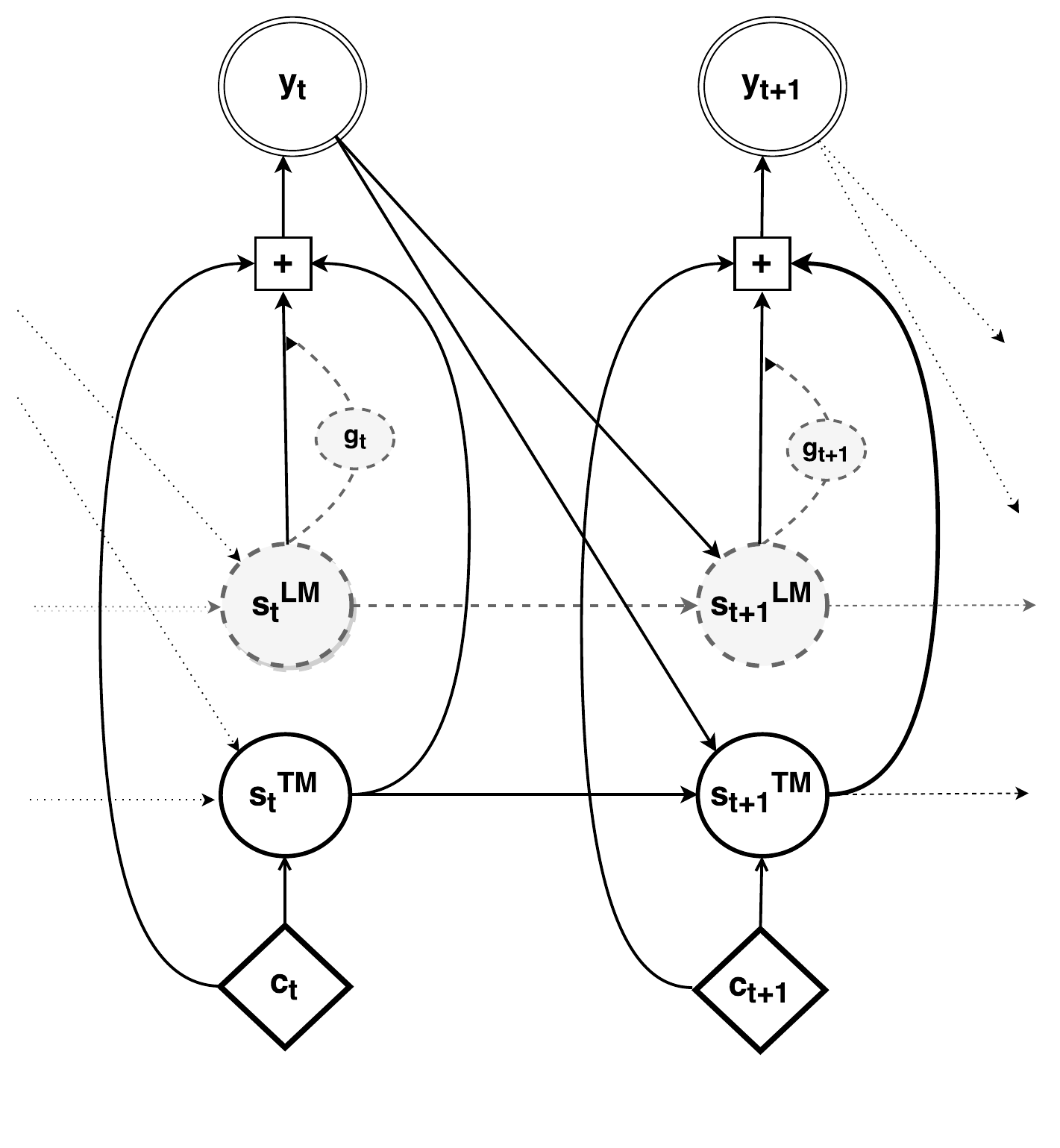}
    \end{minipage}

\begin{minipage}{0.48\textwidth}
        \centering
            (a) Shallow Fusion (Sec.~\ref{sec:shallow})
\end{minipage}
\hspace{2mm}
    \begin{minipage}{0.48\textwidth}
        \centering
            (b) Deep Fusion (Sec.~\ref{sec:deep})
    \end{minipage}
    \caption{Graphical illustrations of the proposed fusion methods.}
\label{fig:fusion}
\end{figure*}

See Fig.~\ref{fig:fusion} (a) for illustration.

\subsection{Deep Fusion}
\label{sec:deep}

In deep fusion, we integrate the RNNLM and the decoder of the NMT by
concatenating their hidden states next to each other (see Fig.~\ref{fig:fusion}
(b)). The model is then finetuned to use the hidden states from both of these
models when computing the output probability of the next word (see
Eq.~\eqref{eq:dec:output}). Unlike the vanilla NMT (without any language model
component), the hidden layer of the deep output takes as input the hidden state
of the RNNLM in addition to that of the NMT, the previous word and the context
such that

\begin{equation}
    \begin{split}
    \label{eq:dec:output_deepfusion}
    p(&\vy_t | \vy_{<t}, \vx) \propto \\
    &\exp(\vy_t^{\top}(\mW_o 
    \f_o(\vs_t^{\text{LM}}, \vs_t^{\text{TM}}, \vy_{t-1}, \vc_t) + \vb_o)),
    \end{split}
\end{equation}

where again we use the superscripts $^{\text{LM}}$ and $^{\text{TM}}$ to
denote the hidden states of the RNNLM and NMT respectively.

During the finetuning of the model, we tune only the parameters that were used
to parameterize the output~\eqref{eq:dec:output_deepfusion}. This is to ensure
that the structure learned by the LM from monolingual corpora is not
overwritten. It is possible to use monolingual corpora as well while finetuning
all the parameters, but in this paper, we alter only the output parameters in
the stage of finetuning.

\subsubsection{Balancing the LM and TM}
\label{sec:controller}

In order for the decoder to flexibly balance the input from the LM and TM, we
augment the decoder with a ``controller'' mechanism. The need to flexibly
balance the signals arises depending on the work being translated. For
instance, in the case of Zh-En, there are no Chinese words that correspond to
articles in English, in which case the LM may be more informative.  On the
other hand, if a noun is to be translated, it may be better to ignore any
signal from the LM, as it may prevent the decoder from choosing the correct
translation.  Intuitively, this mechanism helps the model dynamically weight
the different models depending on the word being translated. 

The controller mechanism is implemented as a function taking the hidden state of
the LM as input and computing
\begin{align}
    \label{eq:controller}
    g_t = \sigma\left( \vv^{\top}_g \vs^{\text{LM}}_{t} + b_g \right),
\end{align}
where $\sigma$ is a logistic sigmoid function. $\vv_g$ and $b_g$ are learned
parameters.

The output of the controller is then multiplied with the hidden state of the LM.
This lets the decoder use the signal from the TM fully, while the controller
controls the magnitude of the LM signal. 

In our experiments, we empirically found that it was better to initialize the
bias $\vb_g$ to a small, negative number. This allows the decoder to decide the
importance of the LM only when it is deemed necessary.~\footnote{
    In all our experiments, we set $\vb_g = -1$ to ensure that $\vg_t$ is
    initially $0.26$ on average.
}

\section{Datasets}
\label{sect:dataset}

We evaluate the proposed approaches on four diverse tasks: Chinese to English
(Zh-En), Turkish to English (Tr-En), German to English (De-En) and Czech to
English (Cs-En). We describe each of these datasets in more detail below.

\subsection{Parallel Corpora}

\subsubsection{Zh-En: OpenMT'15}

We use the parallel corpora made available as a part of the NIST OpenMT'15
Challenge. Sentence-aligned pairs from three domains are combined to form a
training set: (1) SMS/CHAT and (2) conversational telephone speech (CTS) from
DARPA BOLT Project, and (3) newsgroups/weblogs from DARPA GALE Project. In
total, the training set consists of 430K sentence pairs (see
Table~\ref{table:stats} for the detailed statistics). We train models with this
training set and the development set (the concatenation of the provided
development and tune sets from the challenge), and evaluate them on the test
set. The domain of the development and test sets is restricted to CTS.

\begin{table*}[t]
    \centering

    \begin{minipage}{0.48\textwidth}
        \centering
    \begin{tabular}{c | c c}
        & Chinese & English \\
        \hline
        \hline
        \# of Sentences & \multicolumn{2}{c}{436K} \\
        \hline
        \# of Unique Words & 21K & 150K \\
        \hline
        \# of Total Words & 8.4M & 5.9M \\
        \hline
        Avg. Length & 19.3 & 13.5 \\
    \end{tabular}

    (a) Zh-En
    \end{minipage}
    \hspace{2mm}
    \begin{minipage}{0.48\textwidth}
        \centering
    \begin{tabular}{c | c c}
        & Turkish & English \\
        \hline
        \hline
        \# of Sentences & \multicolumn{2}{c}{160K} \\
        \hline
        \# of Unique Words & 96K$^\star$ & 95K \\
        \hline
        \# of Total Words & 11.4M$^\star$ & 8.1M \\
        \hline
        Avg. Length & 31.6 & 22.6 \\
    \end{tabular}

    (b) Tr-En
    \end{minipage}
    \vspace{3mm}
    \begin{minipage}{0.48\textwidth}
        \centering 
    \begin{tabular}{c | c c} 
        & Czech & English \\
        \hline
        \hline
        \# of Sentences & \multicolumn{2}{c}{12.1M} \\
        \hline
        \# of Unique Words & 1.5M & 911K \\
        \hline
        \# of Total Words & 151M & 172M \\
        \hline
        Avg. Length & 12.5 & 14.2 \\ 
    \end{tabular}

    (c) Cs-En
    \end{minipage}
    \hspace{1mm}
    \begin{minipage}{0.48\textwidth}
        \centering 
    \begin{tabular}{c | c c} 
        & German &English \\
        \hline
        \hline
        \# of Sentences & \multicolumn{2}{c}{4.1M} \\
        \hline
        \# of Unique Words & 1.16M$^\dagger$ & 742K \\
        \hline
        \# of Total Words & 11.4M$^\dagger$ & 8.1M \\
        \hline
        Avg. Length & 24.2 & 25.1 \\ 
    \end{tabular}
    (d) De-En
    \end{minipage}
    \caption{Statistics of the Parallel Corpora. $\star$: After segmentation,
    $\dagger$: After compound splitting.}
\label{table:stats}
\end{table*}

\paragraph{Preprocessing}
Importantly, we did ``not segment'' the Chinese sentences and considered each
character as a symbol, unlike other approaches which use a separate
segmentation tool to segment the Chinese characters into
words~\cite{Devlin2014}. Any consecutive non-Chinese characters such as Latin
alphabets were, however, considered as an individual word. Lastly, we removed
any HTML/XML tags from the corpus, chose only the intended meaning word if both
intended and literal translations are available, and ignored any indicator of,
e.g., typos. The only preprocessing we did on the English side of the corpus
was a simple tokenization using the tokenizer from Moses.~\footnote{\url{https://github.com/moses-smt/mosesdecoder/blob/master/scripts/tokenizer/tokenizer.perl}}.

\subsubsection{Tr-En: IWSLT'14}

We used the WIT parallel corpus~\cite{cettolo2012} and SETimes parallel corpus
made available as a part of IWSLT'14 (machine translation track). The
corpus consists of the sentence-aligned subtitles of TED and TEDx talks, and we
concatenated dev2010 and tst2010 to form a development set, and tst2011,
tst2012, tst2013 and tst2014 to form a test set. See Table~\ref{table:stats}
for the detailed statistics of the parallel corpora.

\paragraph{Preprocessing}
As done with the case of Zh-En, initially we removed all special symbols from the 
corpora and tokenized the Turkish side with the tokenizer provided by
Moses. To overcome the exploding vocabulary due to the rich inflections 
and derivations in Turkish, we segmented each Turkish sentence into a sequence 
of sub-word units using Zemberek\footnote{\url{https://github.com/ahmetaa/zemberek-nlp}} 
followed by morphological disambiguation on the morphological analysis~\cite{sak2007}. We
removed any non-surface morphemes corresponding to, for instance, part-of-speech
tags. 

\subsection{Cs-En and De-En: WMT'15}
For the training of our models, we used all the  available training
data provided for Cs-En and De-En in the WMT'15 competition. We used newstest2013
as a development set and newstest2014 for a test set.  The detailed statistics of the
parallel corpora is provided in Table \ref{table:stats}.

\paragraph{Preprocessing}
We tokenized the datasets with Moses tokenizer first. Sentences longer than
eighty words and those that have large mismatch between lengths of the source and
target sentences were removed from the training set. Then, we filtered the
training data by removing sentence pairs in which one sentence (or both) was
written in the wrong language by using a language detection toolkit~\cite{nakatani2010langdetect}, 
unless the sentence had 5 words or less. For De-En, we also split the compounds in the
German side by using Moses. Finally we shuffled the training corpora seven
times and concatenated its outputs.

\subsection{Monolingual Corpora}

The English Gigaword corpus by the Linguistic Data Consortium (LDC), which mainly 
consists of newswire documents, was allowed in both OpenMT'15 and IWSLT-15 
challenges for language modelling. We used the tokenized Gigaword corpus
without any other preprocessing step to train three different RNNLM's to fuse into NMT 
for Zh-En, Tr-En and the WMT'15 translation tasks (De-En and Cs-En.)


\section{Settings}
\label{sec:experiments}

\subsection{Training Procedure}
\label{sect:modelDetails}

\subsubsection{Neural Machine Translation}

The input and output of the network were sequences of one-hot vectors whose
dimensionality correspond to the sizes of the source and target vocabularies,
respectively. We constructed the vocabularies with the most common words in the
parallel corpora. The sizes of the vocabularies for Chinese, Turkish and English were
10K, 30K and 40K, respectively, for the Tr-En and Zh-En tasks. Each word was
projected into the continuous space of $620$-dimensional Euclidean space first
to reduce the dimensionality, on both the encoder and the decoder.  We chose
the size of the recurrent units for Zh-En and Tr-En to be $1,200$ and $1,000$ respectively.

In Cs-En and De-En experiments, we were able to use larger vocabularies. We trained our NMT model for 
Cs-En and De-En with large vocabularies using the importance sampling based technique
introduced in \cite{Jean2014} and with this technique we were able to use large vocabulary 
of size $200k$.


Each model was optimized using Adadelta~\cite{Zeiler2012} with minibatches of
$80$ sentence pairs. At each update, we normalized the gradient such that if the $L_2$
norm of the gradient exceeds $5$, gradient is renormalized back to
$5$~\cite{Pascanu2013}. For the non-recurrent layers (see
Eq.~\eqref{eq:dec:output}), we used dropout \cite{hinton2012improving} and additive Gaussian noise 
(mean $0$ and std. dev. $0.001$) on each parameter to prevent overfitting~\cite{graves2011practical}. 
Training was early-stopped to maximize the performance on the development set measured by
BLEU.~\footnote{We compute the BLEU score using the multi-blue.perl script from
Moses on tokenized sentence pairs.} We initialized all recurrent weight
matrices as random orthonormal matrices. 

\subsubsection{Language Model}

We trained three RNNLM's with $2,400$ long short-term memory (LSTM) \cite{hochreiter1997long}
units on English Gigaword Corpus using respectively the vocabularies 
constructed separately from the English sides of Zh-En and Tr-En corpora. 
The third language model was trained using $2,000$ LSTM units on the English 
Gigaword Corpus again but with a vocabulary constructed from the intersection the 
English sides of Cs-En and De-En. The parameters of the former two language models were
optimized using RMSProp~\cite{tieleman2012lecture}, and Adam optimizer~\cite{Kingma2014} was used
for the latter one. Any sentence with more than ten percent of its
words out of vocabulary was discarded from the training set.  We
did early-stopping using the perplexity of development set.

\subsection{Shallow and Deep Fusion}
\label{sect:fusion_details}

\subsubsection{Shallow Fusion}
The hyperparameter $\beta$ in Eq.~\eqref{eq:dec:output_shallowfusion} was selected
to maximize the translation performance on the development set, from the range
$0.001$ and $0.1$. In preliminary experiments, we found it important to
renormalize the softmax of the LM without the end-of-sequence and out-of-vocabulary 
symbols ($\beta=0$ in Eq. ~\eqref{eq:dec:output_shallowfusion}). 
This may be due to the difference in the domains of TM and LM.

\subsubsection{Deep Fusion}

We finetuned the parameters of the deep output layer
(Eq.~\eqref{eq:dec:output_deepfusion}) as well as the controller (see
Eq.~\eqref{eq:controller} using the Adam optimizer for Zh-En,
and RMSProp with momentum for Tr-En. During the finetuning, the dropout
probability and the standard deviation of the weight noise were set to $0.56$ and
$0.005$, respectively. Based on our preliminary experiments,
we reduced the level of regularization after the first $10K$ updates. In Cs-En and De-En tasks with 
large vocabularies, the model parameters were finetuned using Adadelta while
scaling down the magnitude of the update steps by $0.01$.

\subsubsection{Handling Rare Words}
On the De-En and Cs-En translation tasks, we replaced the unknown words generated 
by the NMT with the words the NMT assigned to which the highest score
in the source sentence (Eq. \eqref{eq:dec:alpha}). We copied the selected source word in the place of the corresponding unknown
token in the target sentence. This method is similar to the technique proposed 
by \cite{luong2014addressing} for addressing rare words. But instead of relying on 
an external alignment tool, we used the attention mechanism of the NMT model to extract alignments.
This method consistently improved the results by approximately $1.0$ BLEU score.

\section{Results and Analysis}

\subsection{Zh-En: OpenMT'15}
\label{sec:zh_en_result}

In addition to NMT-based systems, we also trained a phrase-based as well as
hierarchical phrase-based SMT systems \cite{koehn2003statistical,Chiang2005hierarchical}
with/without re-scoring by an external neural language
model~(CSLM)~\cite{schwenk2007continuous}. We present the results in
Table~\ref{tab:zh_en_openmt2014}. 

We observed that integrating an additional LM by deep fusion (see
Sec.~\ref{sec:deep}) helped the models achieving better performance in general,
except in the case of the CTS task. We noticed that the NMT-based models,
regardless of whether the LM was integrated or not, outperformed the more
traditional phrase-based SMT systems. 

\begin{table}[h]
    \centering
\begin{tabular}{l|cc|cc}
 & \multicolumn{2}{c|}{SMS/CHAT} & \multicolumn{2}{c}{CTS} \\ 
 & Dev           & Test         & Dev        & Test       \\ 
 \hline 
 \hline
PB            & 15.5          & 14.73        & 21.94      & 21.68      \\
+ CSLM     & 16.02         & 15.25        & 23.05      & 22.79      \\ 
\hline
HPB           & 15.33             & 14.71            & 21.45      & 21.43      \\
+ CSLM    & 15.93             & 15.8            & 22.61      & 22.17      \\ 
\hline
\hline
NMT                  & 17.32         & 17.36        & 23.4       & \textbf{23.59}      \\
Shallow & 16.59         & 16.42        & 22.7       & 22.83      \\ 
Deep & \textbf{17.58}& \textbf{17.64}& \textbf{23.78}      & 23.5       \\ 
\end{tabular}
\caption{Results on the task of Zh-En. PB and HPB stand for the phrase-based and hierarchical
phrase-based SMT systems, respectively.}
\label{tab:zh_en_openmt2014}
\end{table}

\begin{table*}[t]
\centering
\begin{tabular}{l|| cc |cccc}
& \multicolumn{2}{c|}{Development Set} & \multicolumn{4}{c}{Test Set}  \\ 
 & \textbf{dev2010} & \textbf{tst2010} & \textbf{tst2011} & \textbf{tst2012} & \textbf{tst2013} & \textbf{Test 2014}  \\ 
 \hline \hline
 \textbf{Previous Best {\scriptsize (Single)}} & 15.33 & 17.14 & 18.77 & 18.62 & 18.88 & - \\ 
 \textbf{Previous Best {\scriptsize (Combination)}} & - & 17.34 & 18.83 & 18.93 & 18.70 & - \\
 \hline
\textbf{NMT} & 14.50 & 18.01 & 18.40 & 18.77 & 19.86 & 18.64 \\
\textbf{NMT+LM (Shallow)} & 14.44 & 17.99 & 18.48 & 18.80 & 19.87 & 18.66 \\
\textbf{NMT+LM (Deep)} & \textbf{15.69} & \textbf{19.34} & \textbf{20.17} & \textbf{20.23} & \textbf{21.34} & \textbf{20.56} \\ 
\end{tabular}
\caption{Results on Tr-En. We show for each set separately to make it easier to
compare to previously reported scores.}
\label{tab:tr_en_iwslt2014}
\end{table*}

\begin{table}[t]
    \centering 
\begin{tabular}{l|cc|cc}
 & \multicolumn{2}{c|}{De-En} & \multicolumn{2}{c}{Cs-En} \\ 
 & Dev           & Test       & Dev     & Test \\
 \hline
 \hline
 NMT Baseline & 25.51 & 23.61 & 21.47 & 21.89 \\
 \hline
 Shallow Fusion & 25.53  & 23.69 &21.95 & 22.18  \\
 Deep Fusion & \textbf{25.88} & \textbf{24.00} & \textbf{22.49} & \textbf{22.36}\\
\end{tabular}
\caption{Results for De-En and Cs-En translation tasks on WMT'15 dataset.}
\label{tab:csde_en_wmt2015}
\end{table}

\subsection{Tr-En: IWSLT'14}

In Table~\ref{tab:tr_en_iwslt2014}, we present our results on Tr-En. Compared to
Zh-En, we saw a greater performance improvement up to +1.19 BLEU points
from the basic NMT to the NMT integrated with the LM under the proposed method of
deep fusion. Furthermore, by incorporating the LM using deep fusion, the NMT
systems were able to outperform the best previously reported
result \cite{yilmaztubitak} by up to $+1.96$ BLEU points on all of the separate 
test sets.

\subsection{Cs-En and De-En: WMT-15}
We provide the results of Cs-En and De-En on Table \ref{tab:csde_en_wmt2015}. Shallow 
fusion achieved $0.09$ and $0.29$ BLEU score improvements respectively on De-En and Cs-En 
over the baseline NMT model. With deep fusion the improvements of $0.39$ and $0.47$ BLEU score
were observed again over the NMT baseline. 

\subsection{Analysis: Effect of Language Model}

The performance improvements we report in this paper reflect a heavy dependency
on the degree of similarity between the domain of monolingual corpora and the
target domain of translation. 

In the case of Zh-En, intuitively, we can tell that the style of writing in both
SMS/CHAT as well as the conversational speech will be different from that of
news articles (which constitutes the majority of the English Gigaword corpus).
Empirically, this is supported by the high perplexity on the development set
with our LM (see the column Zh-En of Table~\ref{table:ppls}).  This explains the
marginal improvement we observed in Sec.~\ref{sec:zh_en_result}.

On the other hand, in the case of Tr-En, the similarity between the domains of
the monolingual corpus and parallel corpora is higher (see the column Tr-En of
Table~\ref{table:ppls}). This led to a significantly larger improvement in
translation performance by integrating the external language model than the case
of Zh-En.  Similarly, we observed the improvement by both shallow and deep
fusion in the case of De-En and Cs-En, where the perplexity on the development
set was much lower. 

Unlike shallow fusion, deep fusion allows a model to selectively incorporate the
information from the additional LM by the controller mechanism from
Sec.~\ref{sec:controller}. Although this controller mechanism works on per-word
basis, it can be expected that if the additional LM models the target domain
better, the controller mechanism will be more frequently active on average, i.e.,
$\E[g_t] \gg 0$. From Table~\ref{table:ppls}, we can see that, on average, the
controller mechanism is most active with De-En and Cs-En, where the additional LM
was able to model the target sentences best. This effectively means that deep
fusion allows the model to be more robust to the domain mismatch between the TM
and LM, thus suggests why deep fusion was more successful than shallow fusion
in the experiments.

\begin{table}[h]
    \centering 
\begin{tabular}{l||c|c|c|c|c}
           & Zh-En                  & Tr-En    & De-En  &  Cs-En        \\ 
           \hline
           \hline
Perplexity & 223.68                  & 163.73  & 78.20  & 78.20 \\
Average $g$ & 0.23                 & 0.12    & 0.28  & 0.31  \\
Std Dev $g$ & 0.0009              & 0.02                 & 0.003  & 0.008 \\
\end{tabular}
\caption{Perplexity of RNNLM's on development sets and the statistics of the controller
gating mechanism $g$.}
\label{table:ppls}
\end{table}

\section{Conclusion and Future Work}

In this paper, we propose and compare two methods for incorporating monolingual
corpora into an existing NMT system. We empirically evaluate these approaches
(shallow fusion and deep fusion) on low-resource En-Tr (TED/TEDx Subtitles),
focused domain for En-Zh (SMS/Chat and conversational speech) and two
high-resource language pairs: Cs-En and De-En. We show that with our approach
on the Tr-En and Zh-En language pairs, the NMT models trained with deep fusion
were able to achieve better results than the existing phrase-based statistical
machine translation systems (up to a $+1.96$ BLEU points on En-Tr). We also
observed up to a $0.47$ BLEU score improvement for high resource language pairs
such as De-En and Cs-En on the datasets provided in WMT'15 competition over our
NMT baseline. This provides an evidence that our method can also improve the
translation performance regardless of the amount of available parallel corpora.

Our analysis also revealed that the performance improvement from incorporating
an external LM was highly dependent on the domain similarity between the
monolingual corpus and the target task. In the case where the domain of the
bilingual and monolingual corpora were similar (De-En, Cs-En), we observed
improvement with both deep and shallow fusion methods. In the case where they
were dissimilar (Zh-En), the improvement using shallow fusion were much
smaller. This trend might also explain why deep fusion, which implements an
adaptive mechanism for modulating information from the integrated LM, works
better than shallow fusion. This analysis also suggests that future work 
on domain adaption of the language model may further improve translations.


%

\bibliography{myrefs}

\begin{thebibliography}{}

\bibitem[\protect\citename{Bahdanau \bgroup et al.\egroup
  }2014]{bahdanau2014neural}
Dzmitry Bahdanau, Kyunghyun Cho, and Yoshua Bengio.
\newblock 2014.
\newblock Neural machine translation by jointly learning to align and
  translate.
\newblock {\em arXiv preprint arXiv:1409.0473}.

\bibitem[\protect\citename{Cettolo \bgroup et al.\egroup }2012]{cettolo2012}
Mauro Cettolo, Christian Girardi, and Marcello Federico.
\newblock 2012.
\newblock Wit3: Web inventory of transcribed and translated talks.
\newblock {\em Proceedings of the 16th Conference of the European Association
  for Machine Translation (EAMT)}, pages 261--268.

\bibitem[\protect\citename{Chiang}2005]{Chiang2005hierarchical}
David Chiang.
\newblock 2005.
\newblock A hierarchical phrase-based model for statistical machine
  translation.
\newblock In {\em Proceedings of the 43rd Annual Meeting on Association for
  Computational Linguistics}, pages 263--270. Association for Computational
  Linguistics.

\bibitem[\protect\citename{Cho \bgroup et al.\egroup }2014]{Cho2014}
Kyunghyun Cho, Bart van Merrienboer, Caglar Gulcehre, Fethi Bougares, Holger
  Schwenk, and Yoshua Bengio.
\newblock 2014.
\newblock Learning phrase representations using {RNN} encoder-decoder for
  statistical machine translation.
\newblock In {\em Proceedings of the Empiricial Methods in Natural Language
  Processing (EMNLP 2014)}, October.
\newblock to appear.

\bibitem[\protect\citename{Devlin \bgroup et al.\egroup }2014]{Devlin2014}
Jacob Devlin, Rabih Zbib, Zhongqiang Huang, Thomas Lamar, Richard Schwartz, and
  John Makhoul.
\newblock 2014.
\newblock Fast and robust neural network joint models for statistical machine
  translation.
\newblock In {\em Association for Computational Linguistics}.

\bibitem[\protect\citename{Goodfellow \bgroup et al.\egroup
  }2013]{Goodfellow2013}
Ian Goodfellow, David Warde-Farley, Mehdi Mirza, Aaron Courville, and Yoshua
  Bengio.
\newblock 2013.
\newblock Maxout networks.
\newblock In {\em Proceedings of The 30th International Conference on Machine
  Learning}, pages 1319--1327.

\bibitem[\protect\citename{Graves}2011]{graves2011practical}
Alex Graves.
\newblock 2011.
\newblock Practical variational inference for neural networks.
\newblock In {\em Advances in Neural Information Processing Systems}, pages
  2348--2356.

\bibitem[\protect\citename{Hinton \bgroup et al.\egroup
  }2012]{hinton2012improving}
Geoffrey~E Hinton, Nitish Srivastava, Alex Krizhevsky, Ilya Sutskever, and
  Ruslan~R Salakhutdinov.
\newblock 2012.
\newblock Improving neural networks by preventing co-adaptation of feature
  detectors.
\newblock {\em arXiv preprint arXiv:1207.0580}.

\bibitem[\protect\citename{Hochreiter and Schmidhuber}1997]{hochreiter1997long}
Sepp Hochreiter and J{\"u}rgen Schmidhuber.
\newblock 1997.
\newblock Long short-term memory.
\newblock {\em Neural computation}, 9(8):1735--1780.

\bibitem[\protect\citename{Jean \bgroup et al.\egroup }2014]{Jean2014}
S{\'e}bastien Jean, Kyunghyun Cho, Roland Memisevic, and Yoshua Bengio.
\newblock 2014.
\newblock On using very large target vocabulary for neural machine translation.
\newblock {\em arXiv preprint arXiv:1412.2007}.

\bibitem[\protect\citename{Kalchbrenner and Blunsom}2013]{Kalchbrenner2013}
Nal Kalchbrenner and Phil Blunsom.
\newblock 2013.
\newblock Recurrent continuous translation models.
\newblock In {\em Proceedings of the ACL Conference on Empirical Methods in
  Natural Language Processing (EMNLP)}, pages 1700--1709. Association for
  Computational Linguistics.

\bibitem[\protect\citename{Kingma and Ba}2014]{Kingma2014}
Diederik~P. Kingma and Jimmy Ba.
\newblock 2014.
\newblock Adam: {A}{ Method for Stochastic Optimization}.
\newblock {\em ar{X}iv:{\tt 1412.6980 [cs.LG]}}, December.

\bibitem[\protect\citename{Koehn \bgroup et al.\egroup
  }2003]{koehn2003statistical}
Philipp Koehn, Franz~Josef Och, and Daniel Marcu.
\newblock 2003.
\newblock Statistical phrase-based translation.
\newblock In {\em Proceedings of the 2003 Conference of the North American
  Chapter of the Association for Computational Linguistics on Human Language
  Technology-Volume 1}, pages 48--54. Association for Computational
  Linguistics.

\bibitem[\protect\citename{Koehn}2010]{koehn2010}
Philipp Koehn.
\newblock 2010.
\newblock {\em Statistical Machine Translation}.
\newblock Cambridge University Press, New York, NY, USA.

\bibitem[\protect\citename{Luong \bgroup et al.\egroup
  }2014]{luong2014addressing}
Thang Luong, Ilya Sutskever, Quoc~V Le, Oriol Vinyals, and Wojciech Zaremba.
\newblock 2014.
\newblock Addressing the rare word problem in neural machine translation.
\newblock {\em arXiv preprint arXiv:1410.8206}.

\bibitem[\protect\citename{Mikolov \bgroup et al.\egroup
  }2011]{mikolov2011rnnlm}
Tomas Mikolov, Stefan Kombrink, Anoop Deoras, Lukar Burget, and Jan Cernocky.
\newblock 2011.
\newblock Rnnlm-recurrent neural network language modeling toolkit.
\newblock In {\em Proc. of the 2011 ASRU Workshop}, pages 196--201.

\bibitem[\protect\citename{Pascanu \bgroup et al.\egroup }2013]{Pascanu2013}
Razvan Pascanu, Tomas Mikolov, and Yoshua Bengio.
\newblock 2013.
\newblock On the difficulty of training recurrent neural networks.
\newblock In {\em Proceedings of the 30th International Conference on Machine
  Learning (ICML 2013)}.

\bibitem[\protect\citename{Pascanu \bgroup et al.\egroup }2014]{Pascanu2014rec}
R.~Pascanu, C.~Gulcehre, K.~Cho, and Y.~Bengio.
\newblock 2014.
\newblock How to construct deep recurrent neural networks.
\newblock In {\em Proceedings of the Second International Conference on
  Learning Representations (ICLR 2014)}, April.

\bibitem[\protect\citename{Sak \bgroup et al.\egroup }2007]{sak2007}
Ha{\c{s}}im Sak, Tunga G{\"u}ng{\"o}r, and Murat Sara{\c{c}}lar.
\newblock 2007.
\newblock Morphological disambiguation of turkish text with perceptron
  algorithm.
\newblock In {\em Computational Linguistics and Intelligent Text Processing},
  pages 107--118. Springer.

\bibitem[\protect\citename{Schuster and Paliwal}1997]{Schuster1997}
Mike Schuster and Kuldip~K Paliwal.
\newblock 1997.
\newblock Bidirectional recurrent neural networks.
\newblock {\em Signal Processing, IEEE Transactions on}, 45(11):2673--2681.

\bibitem[\protect\citename{Schwenk}2007a]{Schwenk2007}
Holger Schwenk.
\newblock 2007a.
\newblock Continuous space language models.
\newblock {\em Comput. Speech Lang.}, 21(3):492--518, July.

\bibitem[\protect\citename{Schwenk}2007b]{schwenk2007continuous}
Holger Schwenk.
\newblock 2007b.
\newblock Continuous space language models.
\newblock {\em Computer Speech \& Language}, 21(3):492--518.

\bibitem[\protect\citename{Schwenk}2012]{Schwenk2012}
Holger Schwenk.
\newblock 2012.
\newblock Continuous space translation models for phrase-based statistical
  machine translation.
\newblock In Martin Kay and Christian Boitet, editors, {\em Proceedings of the
  24th International Conference on Computational Linguistics (COLIN)}, pages
  1071--1080. Indian Institute of Technology Bombay.

\bibitem[\protect\citename{Shuyo}2010]{nakatani2010langdetect}
Nakatani Shuyo.
\newblock 2010.
\newblock Language detection library for java.

\bibitem[\protect\citename{Sutskever \bgroup et al.\egroup
  }2014]{Sutskever2014}
Ilya Sutskever, Oriol Vinyals, and Quoc Le.
\newblock 2014.
\newblock Sequence to sequence learning with neural networks.
\newblock In {\em Advances in Neural Information Processing Systems (NIPS
  2014)}, December.

\bibitem[\protect\citename{Tieleman and Hinton}2012]{tieleman2012lecture}
Tijmen Tieleman and Geoffrey Hinton.
\newblock 2012.
\newblock Lecture 6.5-rmsprop: Divide the gradient by a running average of its
  recent magnitude.
\newblock {\em COURSERA: Neural Networks for Machine Learning}, 4.

\bibitem[\protect\citename{Y{\i}lmaz \bgroup et al.\egroup
  }2013]{yilmaztubitak}
Ertugrul Y{\i}lmaz, Ilknur~Durgar El-Kahlout, Burak Ayd{\i}n, Zisan~S{\i}la
  Ozil, and Coskun Mermer.
\newblock 2013.
\newblock Tubitak turkish-english submissions for iwslt 2013.
\newblock {\em Proceedings of the 10th International Workshop on Spoken
  Language Translation (IWSLT)}, pages 152--159.

\bibitem[\protect\citename{Zeiler}2012]{Zeiler2012}
Matthew~D Zeiler.
\newblock 2012.
\newblock {ADADELTA}: An adaptive learning rate method.
\newblock {\em ar{X}iv:{\tt 1212.5701 [cs.LG]}}.

\end{thebibliography}
\bibliographystyle{acl}

\end{document}